\pdfoutput=1

\documentclass[11pt]{article}

\usepackage{ACL2023}

\usepackage{times}
\usepackage{latexsym}
\usepackage{amsmath}
\usepackage{longtable}
\usepackage{xtab}
\usepackage{booktabs}
\usepackage{multirow}
\usepackage{ltxtable}
\usepackage{tcolorbox}
\usepackage{etoolbox}
\usepackage{multirow}
\usepackage{algorithm}
\usepackage{algorithmicx}
\usepackage{algpseudocode}
\usepackage{tcolorbox}
\usepackage{amssymb}

\usepackage{float}

\usepackage[T1]{fontenc}

\usepackage[utf8]{inputenc}

\usepackage{microtype}

\usepackage{inconsolata}

\title{Efficient Knowledge Infusion via KG-LLM Alignment}

\author{Zhouyu Jiang\thanks{~~Both authors contributed equally to this work.}\ , Ling Zhong\footnotemark[1]\ , Mengshu Sun\ , Jun Xu\ , Rui Sun\ ,  
\\
\bf{Hui Cai}\ , \bf{Shuhan Luo}\ , \bf{Zhiqiang Zhang}\\ 
Ant Group \\
    \texttt{\{jiangzhouyu.jzy,zhongling.zl,mengshu.sms,lingyao.zzq\}@antgroup.com}}
\begin{document}
\maketitle
\begin{abstract}

To tackle the problem of domain-specific knowledge scarcity within large language models (LLMs), knowledge graph-retrieval-augmented method has been proven to be an effective and efficient technique for knowledge infusion. However, existing approaches face two primary challenges: knowledge mismatch between public available knowledge graphs and the specific domain of the task at hand, and poor information compliance of LLMs with knowledge graphs. In this paper, we leverage a small set of labeled samples and a large-scale corpus to efficiently construct domain-specific knowledge graphs by an LLM, addressing the issue of knowledge mismatch. Additionally, we propose a three-stage KG-LLM alignment strategy to enhance the LLM's capability to utilize information from knowledge graphs. We conduct experiments with a limited-sample setting on two biomedical question-answering datasets, and the results demonstrate that our approach outperforms existing baselines.
\end{abstract}

\section{Introduction}

Recent advancements in large language models (LLMs), such as ChatGPT, have demonstrated impressive capabilities in general-purpose content creation~\cite{chatgpt2022,touvron2023llama}. Nevertheless, their proficiency in domain-specific applications, particularly in the medical field, is notably constrained by insufficient knowledge~\cite{bao2023discmedllm, zhang2023alpacareinstructiontuned, han2023medalpaca}. To enhance the domain-specific performance of LLMs, the primary strategies for knowledge infusion include two main approaches: continual pre-training on domain-specific corpora and retrieval-augmented method, which involves integrating external information into the models.

Compared to continual pre-training, the retrieval-augmented approach is gaining popularity in knowledge-intensive scenarios due to its cost efficiency and enhanced traceability~\cite{lewis2020retrieval, lan2023copy}. Some retrieval-augmented methods involve integrating LLMs with resources directly such as professional literature, news articles and tables through supervised fine-tuning~\cite{borgeaud2022improving,hu2023chatdb}. However, the knowledge required by the model may be scattered among vast amounts of data, and directly retrieving from raw data instances will introduce noise inevitably, preventing the model from effectively utilizing the information. To mitigate this issue, leveraging structured knowledge, especially knowledge graphs (KGs), is an effective method~\cite{moiseev-etal-2022-skill, Ranade2023FABULAIR,wang2023knowledgpt}.

However, the existing KG-retrieval-augmented methods still encounter two principal challenges. The first challenge pertains to knowledge mismatch. While many existing strategies utilize publicly available KGs for knowledge infusion, the knowledge demanded by domain-specific tasks is frequently of a highly specialized nature, leading to a substantial likelihood that the KG might not cover all the requisite information, or might even present gaps. The second challenge involves poor information compliance. The structured format of triples in KGs diverges from the free-flowing format of natural language~\cite{li-etal-2021-shot-knowledge, ke-etal-2021-jointgt} and the target text often includes additional information that is not found in the triples. This disparity can lead to confusion within LLMs, which could result in outputs from the trained model that do not align with the information incorporated from the KG, particularly in scenarios with a scarcity of supervised examples.

In this work, we construct a domain-specific corpus-based knowledge graph efficiently by LLMs and develop a knowledge infusion approach to enhance the ability of LLMs to utilize graph information, enabling them to generate comprehensive, logical, and low-hallucination responses. Firstly, we train a knowledge extraction model based on an LLM using a small amount of labeled data. Then, we obtain a domain knowledge graph that resolves knowledge mismatch by performing extraction on unsupervised domain-specific corpora and reducing errors in the results through simple post-processing. Subsequently, we propose a novel three-phase KG-LLM alignment framework to optimize the exploitation of KG content by LLMs. The framework consists of the following stages:
\begin{itemize}
    \item 
    In the initial pre-learning phase, we synthesize substantial triples-to-text generation task examples derived from the previously mentioned extraction outcomes. We then train a Low-Rank Adapter (LoRA)~\cite{hu2022lora}, designated as K-LoRA, to assimilate the process of KG infusion and acquire proficiency in the domain-specific linguistic modality.
    \item 
    The subsequent phase involves supervised fine-tuning. For each question-answer pair in the training set, we retrieve knowledge graph based on the question, concatenate the resultant subgraph into the input and proceed to train an additional LoRA. This process is designed to refine the model's output, tailoring it to the specific demands of the given task.
    \item 
    The final phase is the alignment with knowledge graph feedback (AKGF). In this phase, we extract knowledge triples from the generated responses and compare them with the KG to provide evaluative feedback on the knowledge correctness. This feedback serves as a basis for further fine-tuning the model to achieve more comprehensive, more logical, and less hallucinatory content.
\end{itemize}

To simulate a realistic context where specialized annotations are scarce, we conduct experiments on limited-sample datasets constructed based on two public biomedical question answering datasets, BioASQ~\cite{Nentidis_2022} and CMedQA~\cite{DBLP:journals/corr/abs-2011-13573}. In summary, our main contributions are as follows:
\begin{itemize}
    \item [1)]
    We propose a modular knowledge infusion framework. Building upon the efficiently constructed KG, our approach aligns LLMs with the KG through lightweight parameter adjustment, addressing issues of knowledge mismatch and poor information compliance. Experimental results demonstrate that our method significantly outperforms the baselines.
    \item[2)] 
    We introduce two innovative strategies, namely "pre-learning" and "AKGF", aiming at forging a stronger link between KGs and LLMs. In pre-learning, we demonstrate that triples-to-text task can serve as a simple and effective knowledge infusion strategy. In AKGF, we illustrate that KGs can function as automated evaluators for the knowledge correctness of generated responses.
\end{itemize}

\section{Related Works}
\label{sec:append-how-prod}
\textbf{Retrieval-augmented LLMs}. Retrieval-augmented generation methods \cite{izacard2020leveraging, lewis2020retrieval, min2022nonparametric, borgeaud2022improving} retrieve relevant information from an external database for the query and enable the LLM to generate results using this information. ChainRAG \cite{xu2023search} focuses on addressing the problem of incorrect knowledge retrieved by information retrieval systems, which can mislead the LLM or disrupt its reasoning chain through their interaction. While these methods enhance factuality, they also introduce new hallucinations. To address this challenge, WebBrain \cite{qian2023webbrain} incorporates both specific information and general knowledge, which are intertwined with text snippets and used as references to complete the task. 


\textbf{LLM-augmented KG Construction}. AutoKG \cite{yu2021autokg} proposes a framework for constructing a KG from unstructured documents using information extraction (IE) and internal semantic alignment. Since the graphs constructed by IE typically suffer from edge sparsity and node redundancy, \citet{wu2023commonsense} have applied contrastive pre-training and node clustering to overcome this issue. Leveraging the capabilities of LLMs, \citet{zhu2023llms} designs prompts for various knowledge graph construction tasks. Another line of research has aimed to extract knowledge from LLMs to construct KGs \cite{bosselut2019comet,hao2022bertnet, west2021symbolic}. Additionally, PiVe \cite{han2023pive} utilizes iterative verification prompts to rectify errors in KGs generated by larger LLMs.

\begin{figure*}[htbp]
\centering
\includegraphics[width=0.85\textwidth]{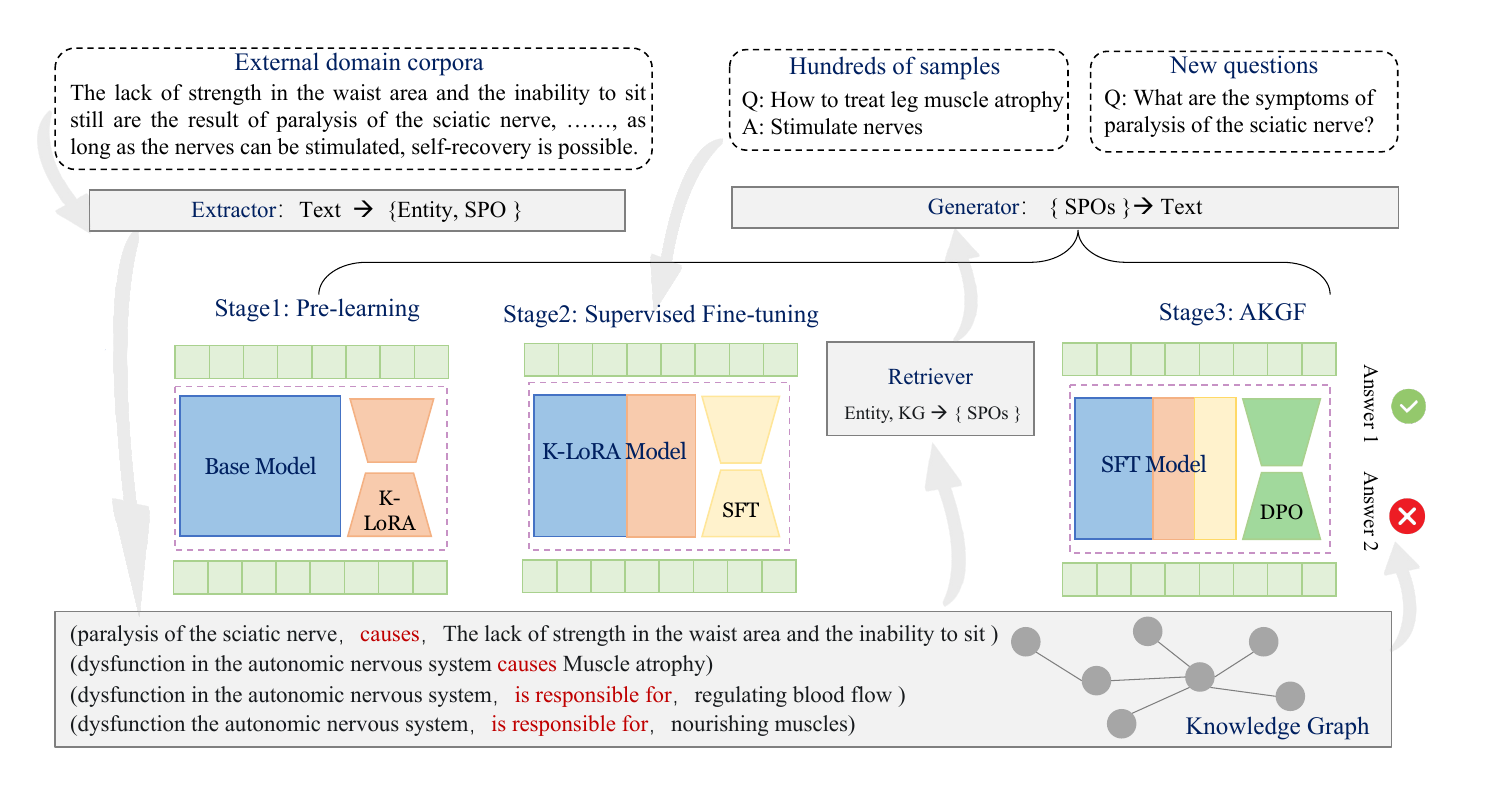} 
\caption{The ELPF framework can be divided into four main stages. 1) \textbf{Efficient construction of domain KGs} The process entails labeling a limited set of examples and developing a LLM-based knowledge extraction system to construct a domain KG from corpora efficiently. 2) \textbf{Pre-learning with K-LoRA}: Gain an understanding of domain-specific knowledge through LoRA-based triples-to-text generation, which is referred to as K-LoRA. 3) \textbf{SFT with KG retrieval}: It involves retrieving subgraphs from the domain-specific KG, modifying the input accordingly and performing supervised fine-tuning. 4) \textbf{AKGF}: The KG acts as an evaluator, providing feedback on knowledge correctness and enabling the model to better align with domain knowledge.}
\label{fig-pipline}
\end{figure*}

\section{Methodology}

Figure \ref{fig-pipline} illustrates the proposed framework, which is referred to as Enhanced LLM with Knowledge Pre-learning and Feedback (ELPF).

\subsection{Efficient construction of domain KGs} \label{construction}
For tasks within a specific domain, publicly available knowledge graphs fail to meet our needs frequently, which is referred to as knowledge mismatch. To counter this issue, a viable solution is to gather a large corpus of domain-specific documents and establish a domain-specific knowledge graph utilizing that corpus. For one unsupervised document $d \in \mathcal{D}$, the process of KG construction can be formalized as Formula \ref{ie}.
\begin{equation}
\begin{aligned}
\{\mathcal{S}^a, \mathcal{P}, O^a\}=\mathcal{F}(d)
\end{aligned}
\label{ie}
\end{equation}

where $\mathcal{F}$ is KG construction system, $\mathcal{S}^a$ is set of subjects, $\mathcal{P}$ is set of defined relationships and $O^a$ is set of objects.

The knowledge triples in the results are organized in the form of Formula \ref{tri}. 
\begin{equation}
\begin{aligned}
\mathcal{T}_{d} = [<s_{1}, p_{1}, o_{11}>,  ..., <s_{j}, p_{k}, o_{jk}>]
\end{aligned}
\label{tri}
\end{equation}
where $o_{jk} = o_{jk1}|o_{jk2}|...|o_{jkl}$. These triples are assembled and merged based on the same subject-relation pair. For example, "\textit{Rome}" and "\textit{Florence}" are both cities of \textit{Italy}, so the instance should be represented as "\textit{<Italy, City, Rome|Florence>}".

However, traditional methods of constructing such graphs can be intricate and often depend on substantial manual labor. Here we have designed an efficient KG construction workflow that requires only minimal annotation, leveraging the advanced semantic comprehension capabilities of LLM. We have streamlined the procedure into three stages:  "knowledge triples extraction", "error removal", and "entity resolution".

Initially, we examine prevalent standards or seek guidance from domain experts to define a schema, which is the set and definitions of entity and relation categories in the domain, and then we manually annotate a small set of examples ($\approx 100$) to generate training data in the format of "text->knowledge triples". Subsequently, we fine-tune an LLM using LoRA, which is a popular parameter-efficient fine-tuning method. Upon merging the trained LoRA parameters into the base model, we perform inference on extensive corpora to derive knowledge triples. Finally, we employ simple post-processing strategies to minimize errors within the extracted triples:

1. Remove results with incorrect output format, such as triples lacking a subject.

2. Remove results where either the subject or object does not appear in the original text.

3. Remove results where the relationship is not in the defined schema.

4. Remove results where the subject and object are the same.

In the entity resolution phase, we utilize an open-source text embedding tool\footnote{https://github.com/shibing624/text2vec \label{em}}  and set a similarity threshold. If the cosine similarity of two subject nodes surpass this threshold, we regard them as equivalent and subsequently combine their respective subgraphs. This merging process contributes to the construction of a comprehensive domain-specific knowledge graph.

We perform a quality assessment on 200 samples of the extracted results from our experimental datasets, where the precision (the ratio of correct triples to the total number of generated triples) surpasses 85\%. In our preliminary experiments, we employed conventional supervised learning methods for extraction (specifically, joint entity and relation extraction based on BERT \cite{Eberts2019SpanbasedJE}, with a sample size greater than 2000) and the eventual evaluation precision was only around 0.8. This is currently a prevalent approach in building knowledge graphs; hence, we consider achieving a precision of approximately 0.85 through post-processing under the current construction workflow to be acceptable. For additional details and statistical outcomes, please refer to Appendix \ref{weakly_ie}.
\subsection{Pre-learning with K-LoRA}

Given that the triple form of KGs deviates from the natural language, LLMs exhibit limited proficiency in processing it. Moreover, acquiring copious amounts of annotated data in specialized domains frequently poses a challenge. Consequently, even with fine-tuning, it remains challenging to enhance the model's capability to leverage information from KGs. We hypothesize that it might be feasible to devise a method for low-cost, extensive data construction that enables the model to assimilate the task format in advance. Fortunately, by inverting the extraction process described earlier, we can create a "triples-to-text" generation task. With extensive fine-tuning on a multitude of instances, the model can be trained to recognize the information format infused by the KG. Additionally, as the target text is domain-specific, the model is able to acquire the unique linguistic style associated with that domain. To boost the fine-tuning process's efficiency, we continue to utilize LoRA-based SFT. We refer to the LoRA obtained in this step as K-LoRA.

\subsection{SFT with KG retrieval} \label{SFT with collaboration}
Pre-learning enables LLMs to better comprehend inputs in the triple form. However, it does not directly resolve specific tasks. Consequently, further refinement through fine-tuning with supervised learning examples remains essential. We adhere to the normal procedure of KG-retrieval-augmented methods~\cite{lewis2020retrieval, 10387715}, which involves retrieving pertinent subgraphs from the previously established domain-specific KG and modifying the input accordingly. The comprehensive input construction is designed to adhere to the following template:
\begin{tcolorbox}
    [colback=gray!20, colframe=gray!100, sharp corners, leftrule={1pt}, rightrule={1pt}, toprule={1pt}, bottomrule={1pt}, left={2pt}, right={2pt}, top={3pt}, bottom={3pt}, label=mybox]
    
    
    \textbf{[KG]:} \{$g_{q}$\}
    
    \vspace{0.2em}
    \textbf{[Instruction]:} Refer to the KG and answer the following question: \{$q$\}
    
\end{tcolorbox}

An initial observation reveals that the subjects and relations inherent in the subgraphs exhibit a significant correlation with the core purpose of the input query. To leverage this observation, we employ an open-source embedding tool\textsuperscript{\ref{em}} to encode all $(s, p)$ pairs within the knowledge graph. Subsequently, we apply the same embedding tool to encode the input query. This approach facilitates the calculation of similarity scores between the query's embedding and those of the top-k $(s, p)$ pairs. Finally, we retrieve the corresponding objects from the original knowledge graph for each $(s, p)$ pair and reconstruct them into triples. These triples are subsequently integrated with the input to provide subgraph information. To maximize the benefits provided by K-LoRA, it is crucial to ensure that the representation of the subgraph remains consistent with the format used during the pre-learning phase.

\subsection{AKGF}
After SFT, the model may still exhibit hallucinations in its responses due to issues such as overfitting. Inspired by the RLHF (Reinforcement Learning with Human Feedback) approach~\cite{ziegler2020finetuning, ouyang2022training}, we hope that the knowledge graph can serve as an automated evaluator, providing feedback on knowledge correctness of the current response, thereby guiding the model towards further optimization.

First, we generate a variety of responses for each query by employing diverse input formats or random seeds. Subsequently, we incorporate the knowledge graph to score and rank these responses. The scoring process entails the utilization of the extraction system described in Section~\ref{construction} to extract triples from these responses, which are then compared with the knowledge graph to ascertain their correctness. The reward is determined by the number of correctly matched knowledge triples. The formula for calculating the reward is represented by Formula \ref{reward}.  
\begin{equation}
\begin{aligned}
reward = \log(r_{spo}+\alpha*r_{e})
\end{aligned}
\label{reward}
\end{equation}
where $\alpha$ is a hyperparameter, $r_{spo}$ represents the number of SPO matches, and $r_{e}$ represents the number of entity matches. For more details on the specific implementation process, please refer to Algorithm~\ref{alg.cps}, where $Jcard$ represents the Jaccard similarity coefficient~\cite{Levandowsky1971DistanceBS}. Appendix \ref{reward_function} demonstrates our automatic reward scoring mechanism using a case example.

To facilitate the training process, we utilize the Direct Preference Optimization (DPO) ~\cite{rafailov2023direct} training strategy, which mitigates sensitivity to reward values, thereby yielding more stable training process. For a comprehensive introduction to the DPO, please refer to Appendix \ref{DPO}. This strategy involves creating pairs of samples according to their reward values. It is crucial to discard any pairs where the difference in rewards is not significant (i.e., $reward_{pos} - reward_{neg} \ge thresh$) and handle issues like repetitive generation in positive samples. To assess the extent of duplication within the positive samples, we can determine the ratio of unique clauses to the overall count of clauses following the deduplication process. Should this ratio fall below a predefined threshold, it would indicate the presence of considerable duplication within the sample, which will be dropped then. By utilizing the knowledge graph for automated evaluation, this method eliminates the requirement of manual scoring, thereby reducing labor costs. Another advantage of this approach is that it is not limited by the quantity of supervised samples, which allows for better learning of knowledge correctness.
\begin{algorithm}
\footnotesize
    \renewcommand{\algorithmicrequire}{\textbf{Input:}}
    \renewcommand{\algorithmicensure}{\textbf{Output:}}
    \newcommand{\var}{\texttt}
    \algnewcommand\And{\textbf{and} }
    \algnewcommand\Or{\textbf{or} }
    \algrenewcommand{\algorithmiccomment}[1]{\hskip0em$//$ #1}
    \caption{Constructing pairwise samples}
    \label{alg1}
    \setlength{\baselineskip}{10.5pt}
    \begin{algorithmic}[1]
        \Require Unsupervised questions $Q$, graph with entities $\mathcal{N}^g$ and SPOs $\{\mathcal{S}^g, \mathcal{P}, O^g\}$
        \For{$q \gets Q$}
            \State $answers = \mathcal{F}(q)$
            \For{$answer \gets answers$}
                \State $\{\mathcal{S}^a, \mathcal{P}, O^a\}=\mathcal{F}_{ie}(answer)$
                \State $r_{spo} \gets 0, r_{e} \gets 0$
                \For{$\{s^{g},p,o^{g}\} \gets \{\mathcal{S}^g, \mathcal{P}, O^g\}$}
                    \If{$Jcard(\{n_s^{a},p,n_o^{a}\},\{s^{g},p,o^{g}\}) \ge thresh_{sim}$}
                        \State $r_{spo} \gets r_{spo} +1$
                    \EndIf
                \EndFor
                \For{$n^{g} \gets \mathcal{N}^g$}
                    \If{$n^{a}=n^{g}$}
                        \State $r_{e} \gets r_{e} +1$
                    \EndIf
                \EndFor
                \State $reward = \log(r_{spo}+\alpha*r_{e})$
            \EndFor
            \For{$ans_{pos}, ans_{neg} \gets answers \times answers$}
                \If{$reward_{pos}-reward_{neg}<thresh$}
                    \State Drop the pairwise sample
                \EndIf
                \If{$ans_{pos}$ contains a lot of repetitive content}
                    \State Drop the pairwise sample
                \EndIf
            \EndFor
        \EndFor
        \Ensure pairwise samples $[Ans_{pos}$, $Ans_{neg}]$
    \end{algorithmic}
\label{alg.cps}
\end{algorithm}

\section{Experimental Settings and Results}

\subsection{Datasets}
We select two biomedical question-answering datasets, CMedQA~\cite{DBLP:journals/corr/abs-2011-13573} and BioASQ~\cite{Nentidis_2022}, for evaluating our model because both demand extensive domain-specific knowledge. CMedQA is a comprehensive dataset of Chinese medical questions and answers, consisting of over 10,000 pairs. In contrast, BioASQ is an English biomedical dataset that includes 4,719 question and answer pairs and 57,360 reference passages. To simulate a scenario with limited samples, we randomly choose 500 instances from each dataset for training and designate 1,000 instances from each for testing. For CMedQA, we employ the answer texts from the non-selected QA pairs as corpora to construct a knowledge graph in a weakly supervised manner. Similarly, with BioASQ, we use all the provided reference passages as the domain-specific corpora.
\subsection{Evaluation Metric}

In our evaluation, we employ multiple metrics, including BLEU (n=4), ROUGE-1, ROUGE-2, and ROUGE-L, to assess the performance of the models. In addition to these automated metrics, we also perform manual evaluations based on five dimensions: fluency, relevance to the question, correctness of the core viewpoint, diversity \& completeness, and knowledge hallucination, using reference answers as a benchmark. Since it is challenging to assign an absolute score through manual evaluation, we sample 200 entries and rank the outputs of models under different settings according to various dimensions. A smaller ranking score indicates better performance, e.g. "1" means the best performance.

\subsection{Experimental Settings}
During the pre-learning stage, we perform fine-tuning of K-LoRA on base LLMs. The learning rate and number of epochs in the pre-learning stage are 5e-5 and 3. During the supervised fine-tuning stage, we establish the similarity threshold for subgraph retrieval to 0.9, and select the top-5 subgraphs. For more hyper-parameters and details, please refer to Appendix~\ref{appendix.implementation_details}.

\renewcommand\arraystretch{1.1}
\begin{table*}[htbp]
    \centering
    \small
    \setlength\aboverulesep{0pt}\setlength\belowrulesep{0pt}
    \begin{tabular*}{\textwidth}{@{\extracolsep{\fill}} l|cccc|cccc}
       \toprule
       \multirow{2}{*}{Model} & \multicolumn{4}{c|}{CMedQA} & \multicolumn{4}{c}{BioASQ} \\
       \cline{2-9}
       & Rouge-1 & Rouge-2 & Rouge-L & BLEU & Rouge-1 & Rouge-2 & Rouge-L & BLEU \\
       \midrule
       ChatGPT-3.5 0-shot  & 18.77 & 2.80 & 14.20 & 1.78 & 27.18 & 9.94 & 21.14 & 5.93\\
       ChatGPT-3.5 2-shot  & 19.61 & 3.14 & 14.66 & 2.53 & 27.45 & 10.26 & 21.42 & 6.11\\
       LLM-base & 19.73 & 3.22 & 14.62 & 1.17 & 13.46 & 2.77 & 8.38 & 1.20 \\
       LLM-base-SFT(No-retrieval) & 17.90 & 2.80 & 14.41 & 2.43 & 27.67 & 11.57 & 23.09 & 7.05 \\ 
       LLM-CP-SFT(No-retrieval) & 18.31 & 2.84 & 14.71 & 2.56 & 26.99 & 11.31 & 23.55 & 7.23 \\ 
       LLM-base-SFT(RAG) & 17.94 & 2.88 & 14.28 & 2.98 & 27.19 & 11.44 & 22.78 & \textbf{9.07}\\
       GAP & 13.23 & 1.488 & 10.23 & 1.82 & 26.5 & 11.31 & \textbf{24.37} & 6.25 \\
       \midrule
       ELPF(ours) & \textbf{19.83} & \textbf{3.86} & \textbf{15.44} & \textbf{3.46} & \textbf{28.55} & \textbf{12.70} & 24.21 & 7.79 \\
       \bottomrule
    \end{tabular*}
    \caption{Performance comparison on CMedQA \& BioASQ. "CP" indicates "continual pre-trained". We consider continual pre-training as a basic method of domain knowledge infusion, on par with other retrieval-based methods. Consequently, we do not report on the outcomes of hybrid approaches.}
    \label{tab:CMedQA_res & BioASQ_res}
\end{table*}

\renewcommand\arraystretch{1.1}
\begin{table*}[htbp]
    \small
    \centering
    \setlength\aboverulesep{0pt}\setlength\belowrulesep{0pt}
    \begin{tabular*}{\textwidth}{@{\extracolsep{\fill}} l|cccc|cccc}
       \toprule
       \multirow{2}{*}{Model} & \multicolumn{4}{c|}{CMedQA} & \multicolumn{4}{c}{BioASQ} \\
       \cline{2-9} 
       & Rouge-1 & Rouge-2 & Rouge-L & BLEU & Rouge-1 & Rouge-2 & Rouge-L & BLEU \\
       \midrule
       ELPF(ours) & \textbf{19.83} & \textbf{3.86} & \textbf{15.44} & \textbf{3.46} & 28.55 & \textbf{12.70} & \textbf{24.21} & \textbf{7.79} \\
       \midrule
       w/o K-LoRA\&AKGF & 18.55 & 3.19 & 14.02 & 2.86 & 28.17 & 11.94 & 23.47 & 7.11 \\
       w/o K-LoRA  & 18.62 & 3.33 & 15.05 & 2.90 & 28.21 & 11.91 & 23.41 & 7.24 \\
       w/o AKGF & 19.77 & 3.85 & 15.31 & 3.35 & \textbf{28.61} & 12.31 & 23.79 & 7.44 \\
       w/o KG retrieval & 19.55 & 3.59 & 15.28 & 3.28 & 28.29 & 11.91 & 23.60 & 7.27 \\
       \bottomrule
    \end{tabular*}
    \caption{Ablation experiment comparison on CMedQA \& BioASQ.}
    \label{tab:Ablation experiment}
\end{table*}

\subsection{Baselines}
\noindent{\textbf{Base LLMs:}} Taking into account the constraints on machine resources and practical use cases, we require models with less than 10B parameters. On the CMedQA dataset, we choose ChatGLM2-6B \cite{zeng2022glm130b} as base model. On the BioASQ dataset, we choose Llama2-chat-7B \cite{touvron2023llama} as base model. Both of the models are initialized with HuggingFace's pre-trained checkpoints\footnote{https://huggingface.co/THUDM/chatglm2-6b}\footnote{https://huggingface.co/meta-llama/Llama-2-7b-chat-hf}. Additionally, we opt to utilize the API of ChatGPT-3.5. For the base LLMs, we present the results of querying the model in a zero-shot scenario. Moreover, to compare the difference with basic continual pre-training method, we conduct continual pre-training on the base LLMs using the aforementioned constructed unsupervised corpus. For the settings of hyper-parameters of continual pre-training, please refer to the Appendix~\ref{appendix.implementation_details}.

\noindent{\textbf{No-retrieval Models:}} We evaluate the performance of base LLMs and continual pre-trained LLMs after LoRA-based SFT with the constructed training set, where the inputs do not contain any retrieval results.

\noindent{\textbf{Retrieval-based Models:}} For KG-level retrieval, we utilize the state-of-the-art KG-to-text method called GAP \cite{colas-etal-2022-gap} as a baseline. GAP enhances KG-to-text generation by incorporating graph-aware elements into pre-trained language models. For document-level retrieval, we compare our approach with the representative method called RAG \cite{lewis2020retrieval}. RAG ensures that the text retrieval source aligns with the unsupervised corpus used for KG construction. The retrieval method employed here is the same as the subgraph retrieval approach discussed in Section \ref{SFT with collaboration}. We place the top-2 retrieved passages on the inputs, then perform LoRA-based SFT and direct query ChatGPT-3.5.

\subsection{Main Results} 
Our results on the CMedQA and BioASQ datasets are shown in Table \ref{tab:CMedQA_res & BioASQ_res}. We observe that the zero-shot querying method achieved ROUGE scores that are close to those obtained through supervised fine-tuning. However, it is worth noting that the zero-shot querying method results in significantly lower BLEU scores on both datasets. These results indicate that the zero-shot querying method does not effectively balance the professionalism and fluency of the generated text. Consequently, this method may not be suitable for generating domain-specific text that meets the desired criteria. 

As for basic 2-shot RAG experiment on ChatGPT-3.5, although there is an improvement compared to the zero-shot baseline for both, the improvement on CMedQA is more pronounced. Case analysis reveals that CMedQA's corresponding corpus is from question-and-answer pairs, whereas BioASQ consists of lengthy paragraphs, which leads to differences in passage format and retrieval quality. This may suggest two things: i. Simple RAG heavily relies on the retriever's capability; ii. Simple RAG is dependent on the format of the text passages. 

In terms of fine-tuning-based methods, our model shows improvements across various metrics. On the CMedQA dataset, our model achieves a 1.03 ROUGE-L improvement and a 1.03 BLEU improvement compared to the vanilla LoRA-based SFT method. On the BioASQ dataset, we have achieved a 1.12 improvement in ROUGE-L and a 0.74 improvement in BLEU. It is worth noting that our method achieves a significant performance improvement even compared to continual pre-training followed by fine-tuning. These results highlight the effectiveness of our proposed KG collaborative method in enhancing the performance of fine-tuning for LLMs. Compared to the GAP method, our approach not only exhibits significant improvements but also offers the advantage of not requiring the full-parameter joint training of a graph encoder with a pre-trained model like GAP. In comparison to RAG, which focuses on document retrieval, our method achieves higher ROUGE scores but lower BLEU scores on the BioASQ dataset. This difference may be attributed to the document retrieval system's ability to recall more extensive information. On the other hand, the process of constructing a KG introduces information loss, which results in ELPF generation relying more on the implicit knowledge of LLM itself when the subgraph is insufficient, leading to lower accuracy. At the same time, document retrieval also introduces more noise, leading to some answers deviating from the original question.

\begin{figure}[htbp]
\centering
\includegraphics[width=7.5cm]{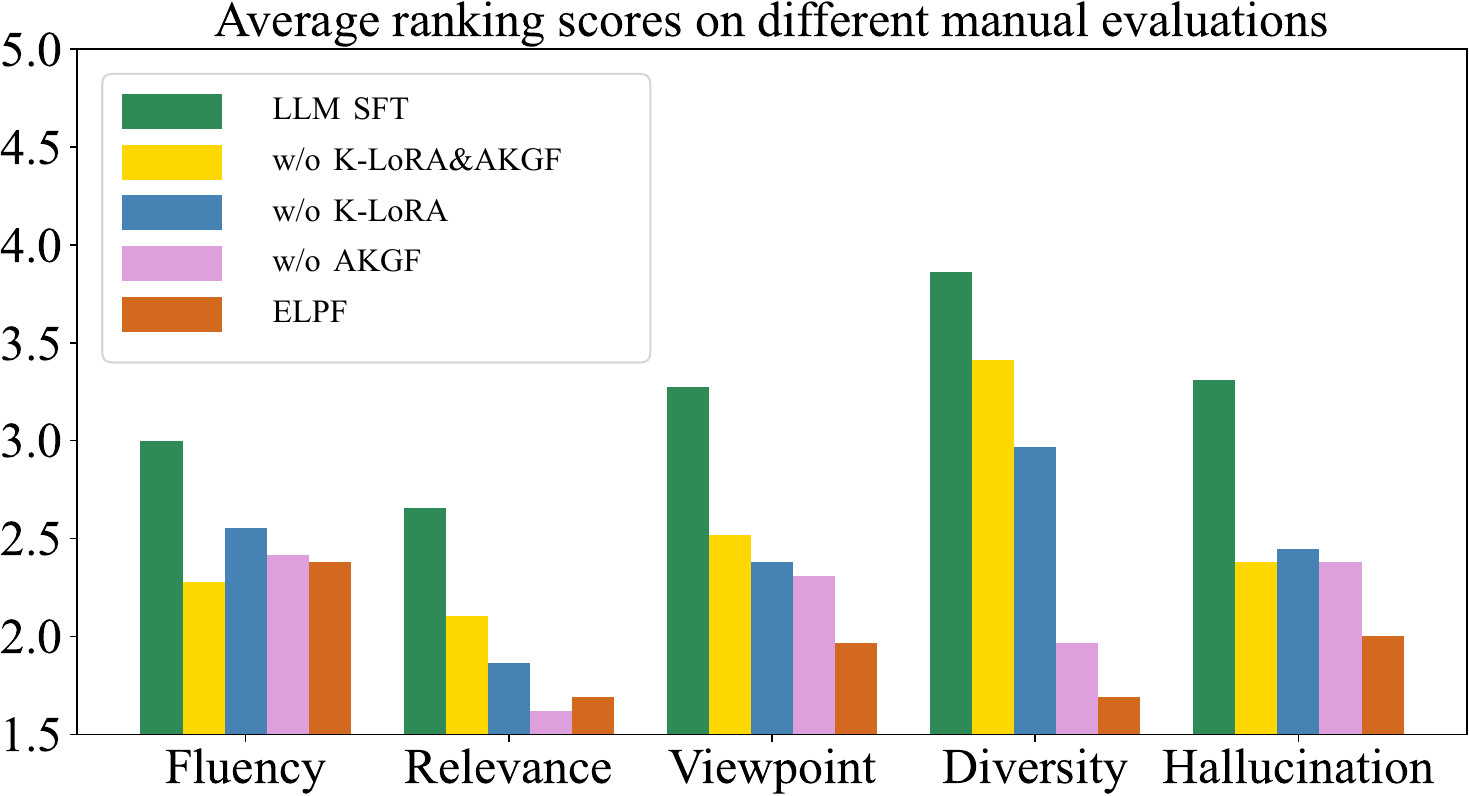}
\caption{On the BioASQ dataset, different methods are ranked based on five human evaluation dimensions: fluency, relevance to the question, correctness of the core viewpoint, diversity \& completeness, and knowledge hallucination. The ranking score represents the manual ranking of the content generated by different models, where a lower ranking score indicates higher quality of the generated content.}
\label{human_eval}
\end{figure}
\section{Analysis}
\subsection{Ablation Studies}
We conduct several ablation experiments to evaluate the effectiveness of each module. These experiments involve the individual removal of K-LoRA, KG prompt, and AKGF, as well as the simultaneous removal of both K-LoRA and AKGF. The results of these experiments, including ROUGE and BLEU scores, can be found in Table \ref{tab:Ablation experiment}. Additionally, the manual evaluation results for BioASQ are presented in Figure \ref{human_eval}. Here are the key observations from our analysis:

(1) Removing K-LoRA leads to the most significant performance drop, reflected in ROUGE, BLEU, and the diversity of knowledge. The main reason is that the format of triples-to-text training samples is similar to the format of the subsequent fine-tuning task, allowing the model to better incorporate the knowledge implied by the input. (2) AKGF has a less significant impact on ROUGE and BLEU metrics. This is because the alignment objective is not focused on replicating the target answer, but rather on incorporating comprehensive, effective, and accurate domain knowledge, even if it is not particularly relevant to the question. It improves the diversity of knowledge, as well as the correctness of viewpoints, and reduces hallucinations, achieving the goal of alignment. (3) The results of manual evaluation indicate that the ablated models with knowledge integration demonstrate improvements over the baseline model that relies solely on fine-tuning, in terms of knowledge correctness (question relevance, viewpoint correctness, and hallucinations) and knowledge diversity. Our ELPF method outperforms others across all dimensions, demonstrating its effectiveness. Appendix~\ref{appendix.case_study} presents a specific case, which allows for a more intuitive understanding of the effectiveness of the answers output by different models.

\subsection{In-depth Analysis of K-LoRA}
To further analyze the overall impact of K-LoRA on the model, we examine its effects on domain awareness and the alignment of generated text with the knowledge graph.
\begin{figure}[htbp]
\begin{center}
\includegraphics[width=7.5cm]{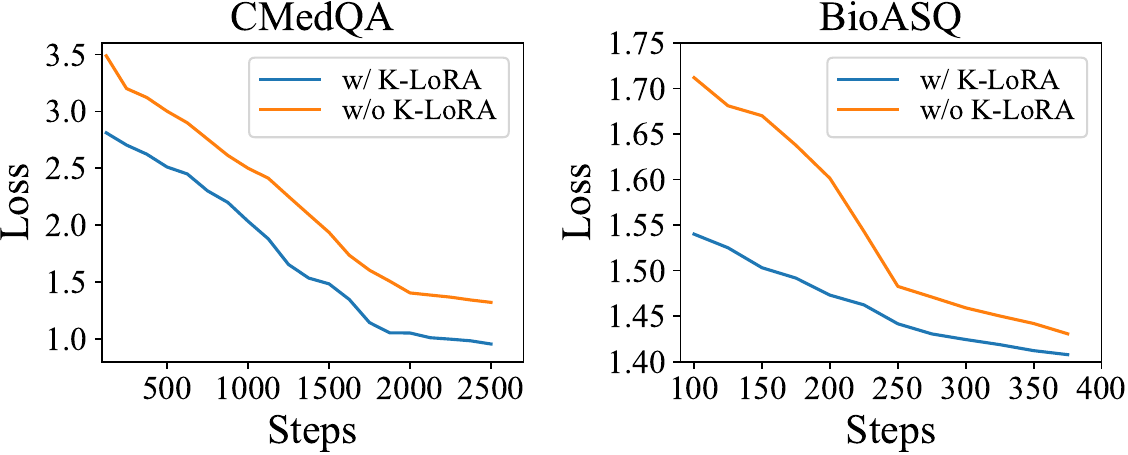} 
\caption{The loss curve of ELPF was compared under the same settings, with and without K-LoRA.}
\label{loss}
\end{center}
\end{figure}
\begin{figure}[htbp]
\begin{center}
\includegraphics[scale=0.6]{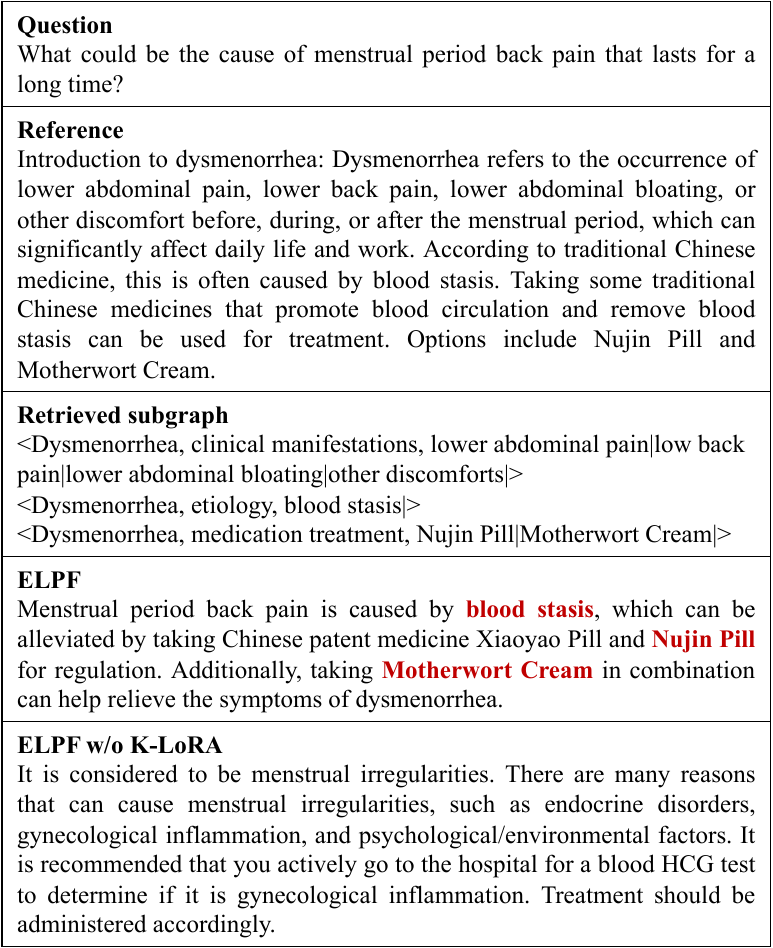} 
\caption{Case study on impact on information compliance of K-LoRA. Information provided by KG in text is highlighted.}
\label{case2}
\end{center}
\end{figure}
K-LoRA aims to enable the LLM to pre-learn within the current domain and generate text with the assistance of the knowledge graph in a weakly supervised manner. Regarding domain awareness, as illustrated in Figure \ref{loss}, we observe a reduction in loss during the model training process and find that the model fused with K-LoRA not only has a smaller initial loss but also converges faster. This indicates that the model develops a certain level of awareness of the current domain through the training and weight fusion of K-LoRA. Therefore, K-LoRA serves as a straightforward and efficient approach for domain learning. In terms of the information compliance of the generated text to the knowledge graph, we analyze the text generated with and without K-LoRA, as shown in Figure \ref{case2}. We notice that although the same knowledge graph information is provided, the original model does not effectively utilize this knowledge graph for generation. On the other hand, the model integrated with K-LoRA relies more on the knowledge graph and generates answers that are closer to the reference answers. This is because the task format of pre-learning and SFT is similar, and K-LoRA enhances the model's ability to adapt to input from the knowledge graph.


\subsection{Knowledge Completeness}
As our approach depends on information from the knowledge graph, this section explores the impact of the knowledge graph's completeness on our method. The completeness of knowledge can be measured by the size and quality of the knowledge graph. First, we explore the influence of graph size. We offer various sizes of KG, including full (100\%), 80\%, 60\%, 40\%, 20\%, and 0\%. The size control is achieved by randomly removing a certain proportion of nodes from the entire graph. Next, we investigate the impact of graph quality. We construct a set of target data to simulate the upper limit of model performance. The target data consists of triples extracted from the reference answers that correspond to the questions. The results are shown in Table \ref{tab:knowledge completeness}. Firstly, we find that reducing the size of the knowledge graph does lead to a decrease in performance, but it is not a purely positive relationship. This is because our knowledge graph contains noise, and the model needs to balance between useful information and noise during the learning process. The model cannot effectively learn when the graph is sparse, resulting in even worse performance compared to not incorporating the graph information. Secondly, we observe that the current results still exhibit a certain gap when compared to the results obtained from the target data. This indicates that there is room for improvement in the quality of the graph constructed by LLMs and the subgraph retrieval method. We will address these issues in future work.

\renewcommand\arraystretch{1.1}
\begin{table}[htbp]
\centering
\small
\setlength\aboverulesep{0pt}\setlength\belowrulesep{0pt}
\begin{tabular}{p{0.8cm}|p{1.2cm}<\centering p{1.2cm}<\centering|p{1.2cm}<\centering p{1.2cm}<\centering}
\toprule
       \multirow{2}{*}{} & \multicolumn{2}{c|}{CMedQA} & \multicolumn{2}{c}{BioASQ} \\
       \cline{2-5}
       & Rouge-L & BLEU & Rouge-L & BLEU \\
       \midrule
       0 \% &15.04 & 2.97& 23.70& 7.23 \\ \hline
       20\%  & 14.98 & 3.02 & 23.59 & 7.14 \\ \hline
       40\% &15.12 & 2.95& 24.20& 7.61\\ \hline
       60\% &15.26 & 3.10& 24.39& 7.70\\ \hline
       80\% &15.30 & 3.22& 24.39& 7.68\\ \hline
       100\% & 15.44 & 3.46 & 24.21 & 7.79\\ \hline
       target &16.40 &3.56 & 25.32& 8.03 \\
       \bottomrule
\end{tabular}
\caption{The performance comparison on knowledge completeness.}
\label{tab:knowledge completeness}
\end{table}


\section{Conclusions}
In this work, we propose a framework for efficiently infuse domain knowledge into LLMs. By employing efficient construction of domain knowledge graphs and a three-stage KG-LLM alignment process, we address the issues of knowledge mismatch and poor information compliance. Experiments demonstrate that our method significantly improves the quality of text generation and knowledge correctness in limited sample scenarios. We hope our work will provide insight into the challenge of connecting KG with LLMs in future exploration.
\section*{Limitations}
Although ELPF is relatively friendly in terms of sample size and computational resources, this method still has certain limitations. Since the construction of the domain knowledge graph is required in both SFT and AKGF, the ELPF method is highly dependent on the quality of the graph construction. However, our graph is established based on weak supervision signals, so there are inevitably noises in the results. Insufficient noise handling can affect the effectiveness of the method. Furthermore, because the self-built domain knowledge graph (KG) is incomplete, it is challenging to detect knowledge errors unless they conflict with known knowledge. Additionally, determining the relevance of the knowledge to the query is a vague concept that is difficult to assess. Therefore, to enhance the stability and versatility of alignment, we have adopted a more conservative strategy in AKGF. This approach somewhat limits the optimization space. However, in actual vertical domain application scenarios, the positive reward or conflict penalty strategies can be adjusted according to the actual situation to achieve better results. Finally, our method focuses on domain-specific text generation. However, due to the limited availability of appropriate public datasets, we only conducted experiments on medical domain texts. This limitation may pose a risk to the generalized ability of our findings in other scenarios.

\bibliography{anthology,custom}
\bibliographystyle{acl_natbib}

\appendix

\section{Weakly Supervised Domain-specific IE System Construction}\label{weakly_ie}
For the annotation standard of the CMedQA dataset, we referred to the CMeIE v2 dataset\footnote{https://tianchi.aliyun.com/dataset/95414}, which is a large-scale Chinese medical domain relation extraction dataset. For BioASQ, we referred to BioRED \cite{10.1093/bib/bbac282}, an English medical relation extraction dataset annotated on the PubMed data source. 

The types of relationship defined in the CMedQA dataset are: ["Differential Diagnosis", "Pathological Typing", "Clinical Manifestation", "Adjuvant Therapy", "Pharmacotherapy", "Surgical Treatment", "Etiology", "Synonyms", "Imaging Examination", "Auxiliary Examination", "Department of Consultation", "Complications", "Laboratory Test", "Susceptible Population", "Genetic Factors", "High-risk Factors", "Pathogenesis", "Site of Onset", "Medical History", "Incidence Rate", "Prognosis", "Age of Onset", "Prevention", "Post-treatment Symptoms", "Pathophysiology", "Transmission Route", "Peak Season", "Histological Examination", "Stage", "Radiotherapy", "Screening", "Chemotherapy", "Risk Assessment Factors", "Metastatic Sites", "Prevalence Area", "Mortality Rate"]. 

The types of relationship defined in the BioASQ dataset are: ["Association", "isa", "Negative\_Correlation", "Positive\_Correlation"].

For each reference dataset, we only utilized its relational schema and manually annotated 100 samples sampled from unsupervised corpora. 

During manual annotation, we assigned two annotators for blind labeling and one quality control personnel for inspection. The final inter-annotator agreement was 0.9, and the accuracy of acceptance was 0.97. During the training, we employed the generative information extraction paradigm and trained a LoRA on top of an LLM. The hyperparameter settings were consistent with those in the SFT stage.

Statistical details of the constructed graph are provided in Table \ref{KG_stats}.
The symbol "\#" denotes a sign for counting. We performed a quality assessment on 200 samples of the extracted results from experimental datasets and calculated the precision (the ratio of correct triples to the total number of generated triples).
\begin{table}[htbp]
\centering
\small
\setlength\aboverulesep{0pt}\setlength\belowrulesep{0pt}
\begin{tabular}{p{1.2cm}|p{1.2cm}<\centering p{1.2cm}<\centering p{1.2cm}}
\toprule
       \cline{2-4}
       Datasets&  \#Subjects & \#Triples & Precision   \\
       \midrule
       CMedQA &25963 & 220111& 0.85\\ \hline
       BioASQ  & 20922 & 53209 & 0.89  \\ \hline
       \bottomrule
\end{tabular}
\caption{Statistics of the constructed domain KGs.}
\label{KG_stats}
\end{table}
\section{Automated Reward Function}\label{reward_function} 
\begin{figure*}[htbp]
\centering
\includegraphics[width=0.8\textwidth]{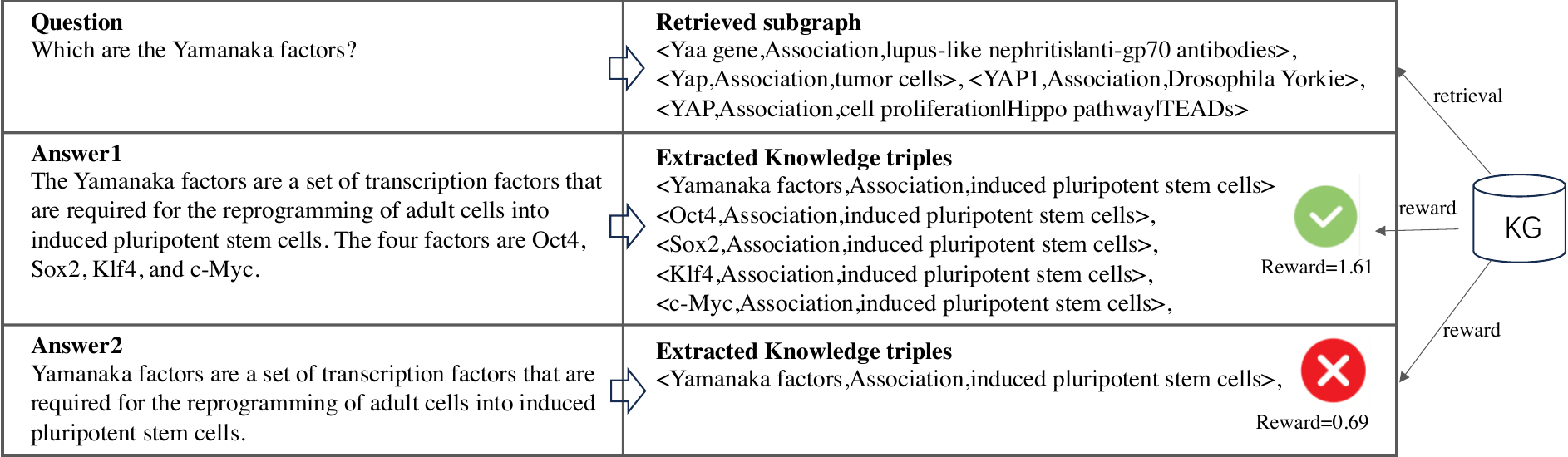} 
\caption{Case study on AKGF dataset generation.}
\label{RLKGF-case}
\end{figure*}

In AKGF, we primarily propose an automated reward scoring mechanism that integrates a Knowledge Graph (KG). Here, we will demonstrate this process through a specific case study as show in Figure \ref{RLKGF-case}. For detailed information about the reward calculation, please refer to Algorithm~\ref{alg.cps}.

\renewcommand\arraystretch{1.1}
\begin{table*}[htbp]
\centering
\small
\setlength\aboverulesep{0pt}\setlength\belowrulesep{0pt}
\begin{tabular}{p{2.5cm}|p{1.2cm}<\centering|p{1.2cm}<\centering|p{1.2cm}<\centering|p{1.2cm}<\centering|p{1.2cm}<\centering|p{1.2cm}<\centering}
\toprule
       \multirow{2}{*}{} & \multicolumn{3}{c|}{CMedQA} & \multicolumn{3}{c}{BioASQ} \\
       \cline{2-7} 
       & K-LoRA & SFT & AKGF & K-LoRA & SFT & AKGF  \\
       \midrule
       LLM &\multicolumn{3}{c|}{ChatGLM2-6B}& \multicolumn{3}{c}{Llama2-chat-7B} \\ \hline
       batch size &32 & 32& 8 &32 & 32& 8 \\ \hline
       fine-tuning type & LoRA & LoRA & LoRA & LoRA & LoRA & LoRA \\ \hline
       epochs &3 & 20& 1 &3 & 3& 1 \\ \hline
       lora rank &8 &8 &8 &8 &8 &8 \\ \hline
       lora target &QKV &QKV &QKV & QKVO & QKVO & QKVO \\ \hline
       learning rate & $5e^{-5}$ & $1e^{-4}$ & $1e^{-6}$ & $5e^{-5}$ & $5e^{-5}$ & $1e^{-6}$ \\ \hline
       max-input-length &512 &512 & 512 &512 &512 &512 \\ \hline
       max-output-length &512 &512 & 512 &512 &512 &512 \\ \hline
       KL-div $\beta$ &- &- & 0.4 & - &- & 0.4 \\ \hline
       top-p & 0.7 & 0.7 & 0.7 & 0.7 & 0.7 & 0.7 \\ \hline
       temperature &0.9 &0.9 &0.9 &0.01 &0.01 &0.01 \\ \hline
       \bottomrule
\end{tabular}
\caption{The parameter settings on CMedQA and BioASQ.}
\label{tab:parameter setting}
\end{table*}

\begin{figure*}[htbp]
\centering
\includegraphics[width=1\textwidth]{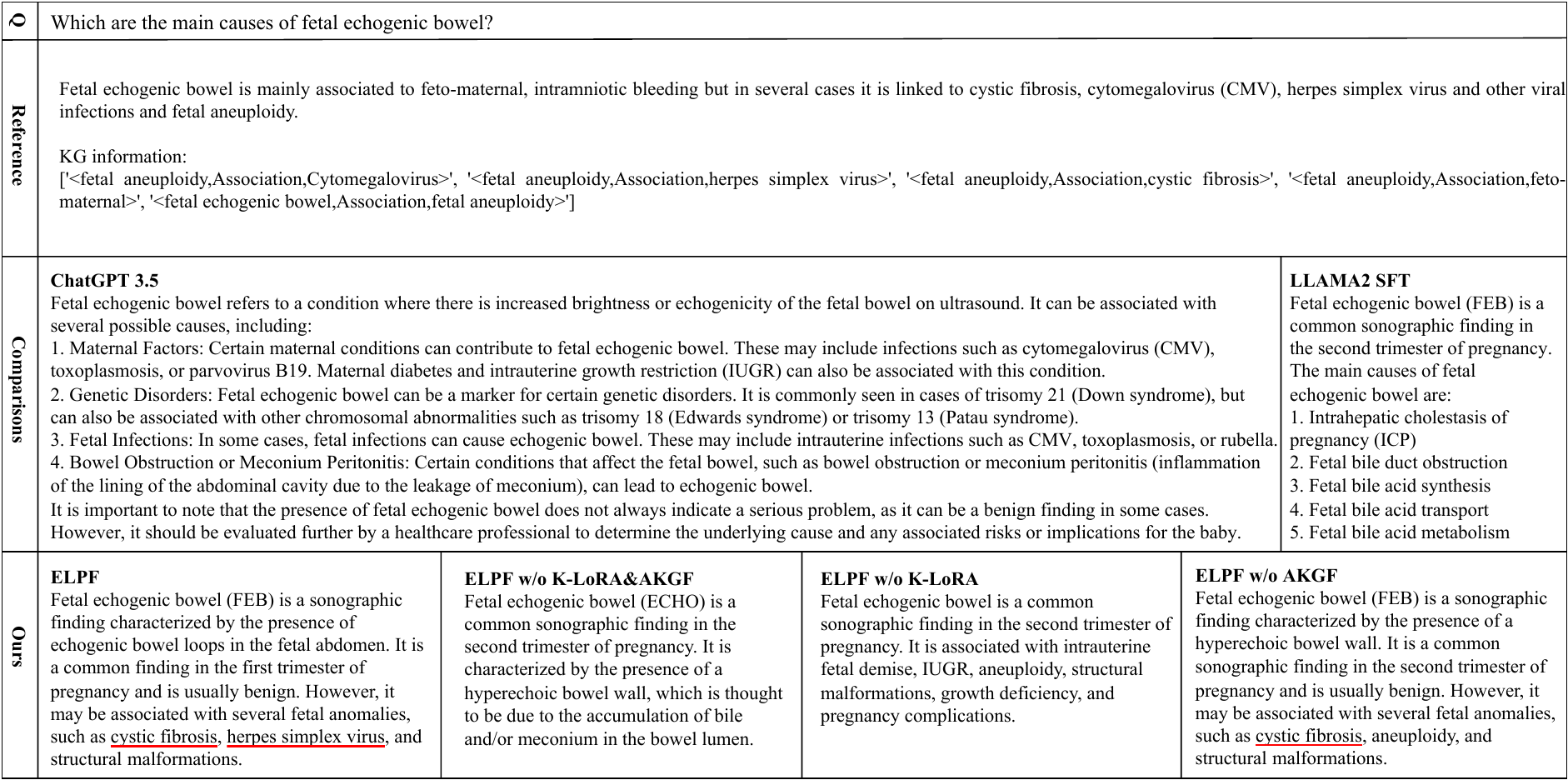} 
\caption{Performance of various models in one case.}
\label{case1}
\end{figure*}

\section{Direct Preference Optimiz ation (DPO)}\label{DPO} 
Construct a static pairwise dataset $\mathcal{D}=\{x^{i},y_{\omega}^{i},y_{l}^{i}\}^{N}_{i=1}$ according to Section 3.3, where $y_{\omega}$ represents the positive samples and $y_{l}$ represents the negative samples, and then perform reward modeling. According to DPO, the reward model $r_{\phi}(x,y)$ is trained using a negative log-likelihood loss as follows:
\begin{equation*}
\begin{aligned}
\mathcal{L} = -\mathbb{E}_{(x,y_{\omega},y_{l})\sim\mathcal{D}}[\log\theta(r_{\phi}(x,y_{\omega})-r_{\phi}(x,y_{l})]
\end{aligned}
\label{dpo}
\end{equation*}
where $\theta$ is the logistic function. In the context of LMs, the network $r_{\phi}(x,y)$ is often initialized from the SFT model $\pi^{SFT}(y|x)$ with the addition of a linear layer on top of the final transformer layer that produces a single scalar prediction for the reward value. To ensure a reward function with lower variance, prior works normalize the rewards, such that $\mathbb{E}_{(x,y)\sim\mathcal{D}}[r_{\phi}(x,y)]=0$ for all $x$.
During the DPO RL phase, use the learned reward function to provide feedback to the language model, with the optimization objective as follows:
\begin{equation*}
\begin{aligned}
\mathcal{J} = \max_{\pi_{\theta}}\mathbb{E}_{x\sim\mathcal{D},y\sim\pi_{\theta}(y|x)}[(r_{\phi}(x,y)]\\-\beta\mathbb{D}_{KL}[\pi_{\theta}(y|x)||\pi_{ref}(y|x)]
\end{aligned}
\label{rl}
\end{equation*}
where $\beta$ is a parameter that controls deviation from the baseline reference policy $\pi_{ref}$, and constraints on the KL divergence ensure that the reward strategy does not deviate too far from the baseline reference strategy (SFT). We also analyzed the impact of the value of the $\beta$ parameter on the training process and selected an optimal parameter for subsequent training, as seen in Table \ref{tab:parameters_select}.
\renewcommand\arraystretch{1.1}
\begin{table}[htbp]
\centering
\small
\setlength\aboverulesep{0pt}\setlength\belowrulesep{0pt}
\begin{tabular}{p{0.8cm}|p{1.2cm}<\centering p{1.2cm}<\centering p{1.2cm}<\centering p{1.2cm}<\centering}
\toprule
       \cline{2-5}
       & Rouge-1 & Rouge-2 & Rouge-L & BLEU \\
       \midrule
       $\beta$=0.1 &28.1 & 11.81& 23.29& 7.2 \\ \hline
       $\beta$=0.2  & 28.2 & 11.88 & 23.36 & 7.25 \\ \hline
       $\beta$=0.4 &28.61 & 12.27& 23.81& 7.42\\ \hline
       \bottomrule
\end{tabular}
\caption{In BioASQ, performance comparison of ELPF on different parameters $\beta$.}
\label{tab:parameters_select}
\end{table}

\section{Implementation Details}
We conduct experiments on four A100 80GB GPUs and two V100 32GB GPUs. For details of the parameters used in the experimental training at each stage, please refer to Table \ref{tab:parameter setting}. As for continual pre-training, we fine-tune full parameter of the LLM with batch\_size=4, epochs=3, learning\_rate=5e-5.
\label{appendix.implementation_details}

\section{Case Study}\label{appendix.case_study}
We evaluate the effectiveness of the model through several case studies, as shown in Figure~\ref{case1}. ELPF provided concise and relatively comprehensive answers regarding the characteristics and main causes of fetal intestinal echoes. It mentioned both physiological and pathological situations. ELPF (w/o AKGF) is close to ELPF in performance. However, the other answers were not as complete. ELPF (w/o K-LoRA\&AKGF) only mentions the physiological condition, while ELPF (w/o K-LoRA) only addresses the pathological factors. Untrained models like ChatGPT-3.5 and Llama2-chat-7B exhibit obvious hallucinations.

\end{document}